\documentclass[journal]{IEEEtran}
\newcommand{\ignore}[1]{}

\usepackage[utf8]{inputenc}
\usepackage[T1]{fontenc}
\usepackage{amsmath,epsfig}
\usepackage{graphicx} 
\usepackage{epstopdf}
\usepackage{url}
\usepackage{multirow}
\usepackage{color}
\usepackage{graphicx}
\usepackage{bm}
\usepackage{amssymb}
\usepackage{cite}
\usepackage{array}
\usepackage[colorlinks,
            linkcolor=red,
            anchorcolor=blue,
            citecolor=green
            ]{hyperref}

\begin{document}
%
\title{Boundary-guided Feature Aggregation Network for Salient Object Detection}
\author{Yunzhi Zhuge, Pingping Zhang, Huchuan Lu}
\maketitle

\markboth{IEEE Signal Processing Letter}{}

\begin{abstract}
Fully convolutional networks (FCN) has significantly improved the performance of many pixel-labeling tasks, such as semantic segmentation and depth estimation.
However, it still remains non-trivial to thoroughly utilize the multi-level convolutional feature maps and boundary information for salient object detection.
In this paper, we propose a novel FCN framework to integrate multi-level convolutional features recurrently with the guidance of object boundary information.
First, a deep convolutional network is used to extract multi-level feature maps and separately aggregate them into multiple resolutions, which can be used to generate coarse saliency maps.
Meanwhile, another boundary information extraction branch is proposed to generate boundary features.
Finally, an attention-based feature fusion module is designed to fuse boundary information into salient regions to achieve accurate boundary inference and semantic enhancement.
The final saliency maps are the combination of the predicted boundary maps and integrated saliency maps, which are more closer to the ground truths.
Experiments and analysis on four large-scale benchmarks verify that our framework achieves new state-of-the-art results.
\end{abstract}


\IEEEpeerreviewmaketitle

\section{Introduction}
\label{sec:intro}
Saliency object detection is a fundamental computer vision task which aims to identify the most eye-catching objects and areas in an image~\cite{Itti1998A}\cite{Achanta2009Frequency}~\cite{Tong2015Salient}~\cite{Wang2015Deep}.
In the past two decades, great success has been made in this pixel-labeling task.
However, due to several inevitable factors such as cluttered backgrounds or blurred boundaries, it still remains a difficult task to combine all hand-tuned cues in an appropriate way.

Recently, deep convolutional neural networks (CNNs) have greatly improved the performances of many computer vision tasks, such as image classification~\cite{Krizhevsky2012ImageNet}, semantic segmentation~\cite{FCN} and visual tracking~\cite{Wang2016STCT,zhang2018non}.
%
%
With the advantages of fully convolutional networks (FCNs)~\cite{FCN}, several FCNs-based attempts have been performed and delivered state-of-the-art performance in predicting saliency maps~\cite{Wang2016Saliency,Lee2016Deep,Liu2016DHSNet}.
Nonetheless, existing models mainly focus on utilizing high-level features extracted from last convolutional layers.
As a result, they are lack of low-level visual information such as object boundary.
Thus, these models tend to predict imperfect results with poorly localized object boundaries.

In this paper, we propose a novel saliency detection method based on multi-level features and boundary cues.
To make full use of the multi-level convolutional features and boundary information, we present a boundary-guided feature aggregating architecture, which simultaneously generates and merges multi-level saliency maps and boundary prediction maps to obtain accurate saliency maps.

Our framework has two streams for saliency prediction.
In the main stream, we predict saliency maps with the incorporated features maps at different resolutions, and these predicted saliency maps are recursively sent to the refinement stage as the inputs.
In another stream, boundary features are extracted through a boundary extraction structure.
To utilize the boundary information, we propose an attention-based feature fusion module to integrate two-stream features.

Our main contributions are as follows:
\begin{itemize}
\item
We propose a boundary-guided feature aggregation network, termed as BFANet, to utilize multi-level convolutional features and boundary cues for salient object detection.
The BFANet first extracts multi-level features, then integrates them into multiple resolutions.
The boundary information fusion is performed to enhance these features to generate finer saliency maps.
\item
We propose a boundary extraction network as a branch of the BFANet, which generates boundary feature maps under the supervision of boundary ground truth.
Besides, we also introduce an attention-based feature fusion module to produce saliency maps with robust boundary.
\item
Extensive experiments on four large-scale benchmarks have shown that our approach performs favorably against other state-of-the-art methods.
\end{itemize}
\section{Our Proposed Method}
\label{sec:previous work}
%
Our method is mainly motivated by the following facts.
First, previous methods usually focus on precisely localizing salient objects, which more or less neglect the sharpness of boundary areas.
To enhance the boundary, we propose a two-steam structure to generate saliency maps with accurate object boundaries.
Secondly, features of variant scales contribute differently to detection.
However, there still exist questions in how to effectively utilize these features.
Therefore, we make a feasible attempt and propose an effective attention-based structure to perform the multi-level feature fusion.

As shown in Fig.~\ref{diagram}, our proposed framework is composed of three components: Aggregating Feature Extraction Network (AFEN), Boundary Prediction Network (BPN) and Attention-based Feature Fusion Module (AFFM).
%
In the following subsections, we will elaborate these components in detail.
\vspace{-2mm}
\subsection{Aggregating Feature Extraction Network}
\begin{figure*}[!t]
\centering
\begin{tabular}{@{}c@{}c}
\includegraphics[width=0.85\linewidth]{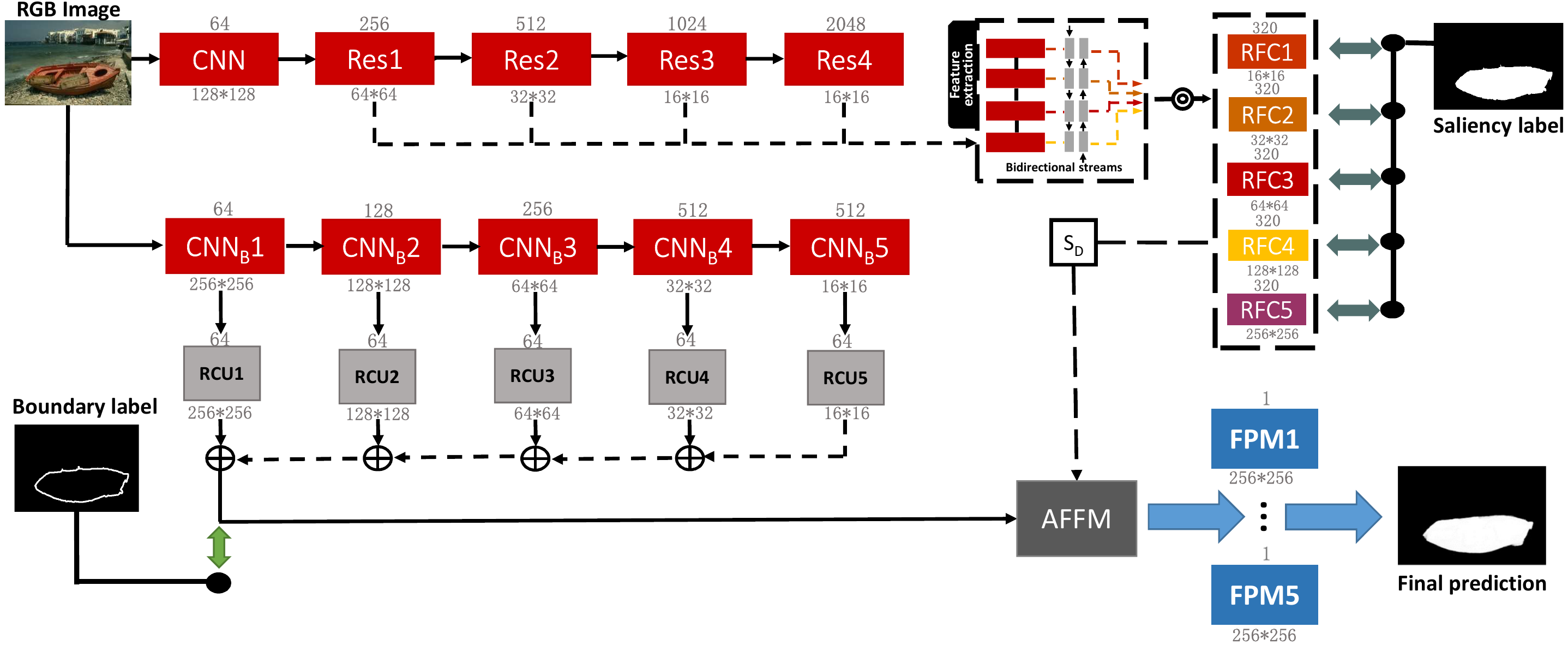} \ &
\includegraphics[width=0.16\linewidth]{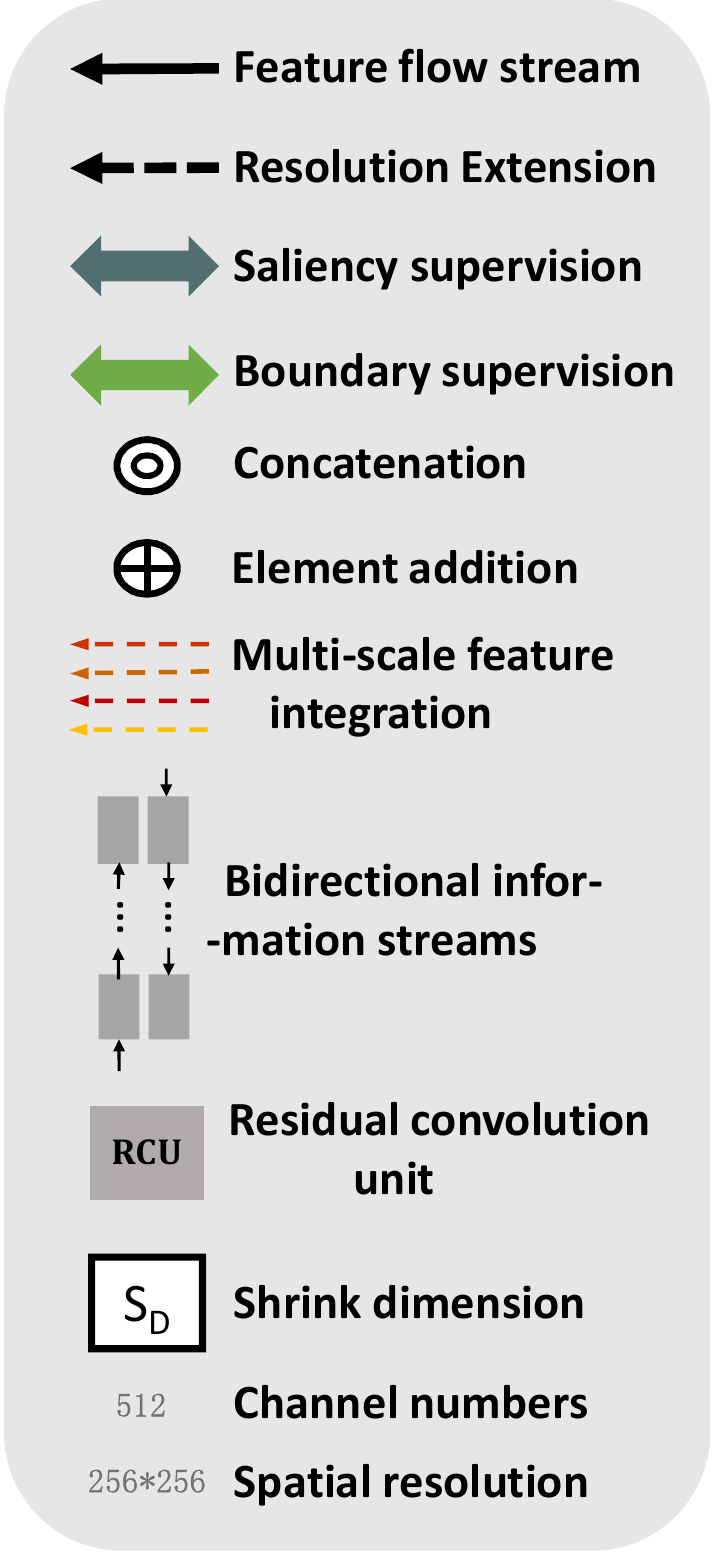} \ \\
\end{tabular}
\caption{
The overall architecture of our proposed BFANet. The upper stream represents the reproduced Amulet~\cite{Zhang2017Amulet}, which is used to extract multi-level convolutional features.
The details of Amulet are described in Section~\ref{sec:AmuletNet for initial prediction}.
In another stream, Boundary Prediction Network (BPN) can progressively enlarge the resolution and refine the details of boundary predictions through a cascaded of residual convolution unit (RCU).
Attention-based feature fusion module (AFFM) is employed to integrate multi-level features and boundary cues.
After that, we obtain multi-level fused feature maps by parallel fused prediction modules (FPM).
The final saliency map is produced with the fused feature maps.
\label{diagram}}
\vspace{-5mm}
\end{figure*}
\label{sec:AmuletNet for initial prediction}
The residual networks (ResNet)~\cite{He2016Deep} have shown excellent performances in many computer vision tasks.
Our feature extraction network is based on the ResNet-101, which extracts multi-level feature maps from raw RGB images for feature integration and saliency prediction.
We modify the ResNet-101 and reduce the resolution of features by a factor of 16 comparing to the input image, as shown in the Fig.~\ref{diagram}.
The feature maps are extracted from specified convolutional layers.
For the balance of resolution, we adopt the resolution-based feature combination structure (RFC)~\cite{Zhang2017Amulet} to integrate multi-level convolutional features.
The RFC structure unifies convolutional features through shrink and extend operations.
More specifically, given an input image $\bf{I}$, the integrated feature maps of scale $\tau\in[1,..,5]$ are computed by
\begin{equation}
\bf{F}^{\tau} = \it{C}_{m=1}^{\text{4}}( \it{R_{m}^{\tau}}(\bf{F}_{\it{m}}(I);\psi_{\it{m}})),
\label{eq:original-bayes}
\end{equation}
where $\it{R_{m}^{\tau}}({\cdot};\psi_{m})$ represents the reshape operator that expands or shrinks the feature maps by a factor of $\psi_{m}$.
$\bf{F}_{\it{m}}$ denotes the $m$-level feature maps.
$\it{C}$ is the concatenation operation in channel-wise.
The resolution of generated feature maps $\bf{F^{\tau}}$ is $[\frac{W}{2^{\tau}},\frac{H}{2^{\tau}}]$.
Besides, to enhance the feature interaction, we adopt the bidirectional information streams~\cite{Zhang2017Amulet}, which integrate multi-level features in both bottom-up and top-down directions.
Although multi-level features have been extracted and fused in this effective way, there still exists a large gap between predicted saliency maps and ground truth.
Due to the defects of down-sampling operations, the predicted saliency maps are blurred or occluded on boundary areas.
\vspace{-2mm}
\subsection{Boundary Prediction Network}
\label{Boundary Prediction Network}
\begin{figure}[!t]
\centering
\begin{tabular}{@{}c}
\includegraphics[width=0.9\linewidth]{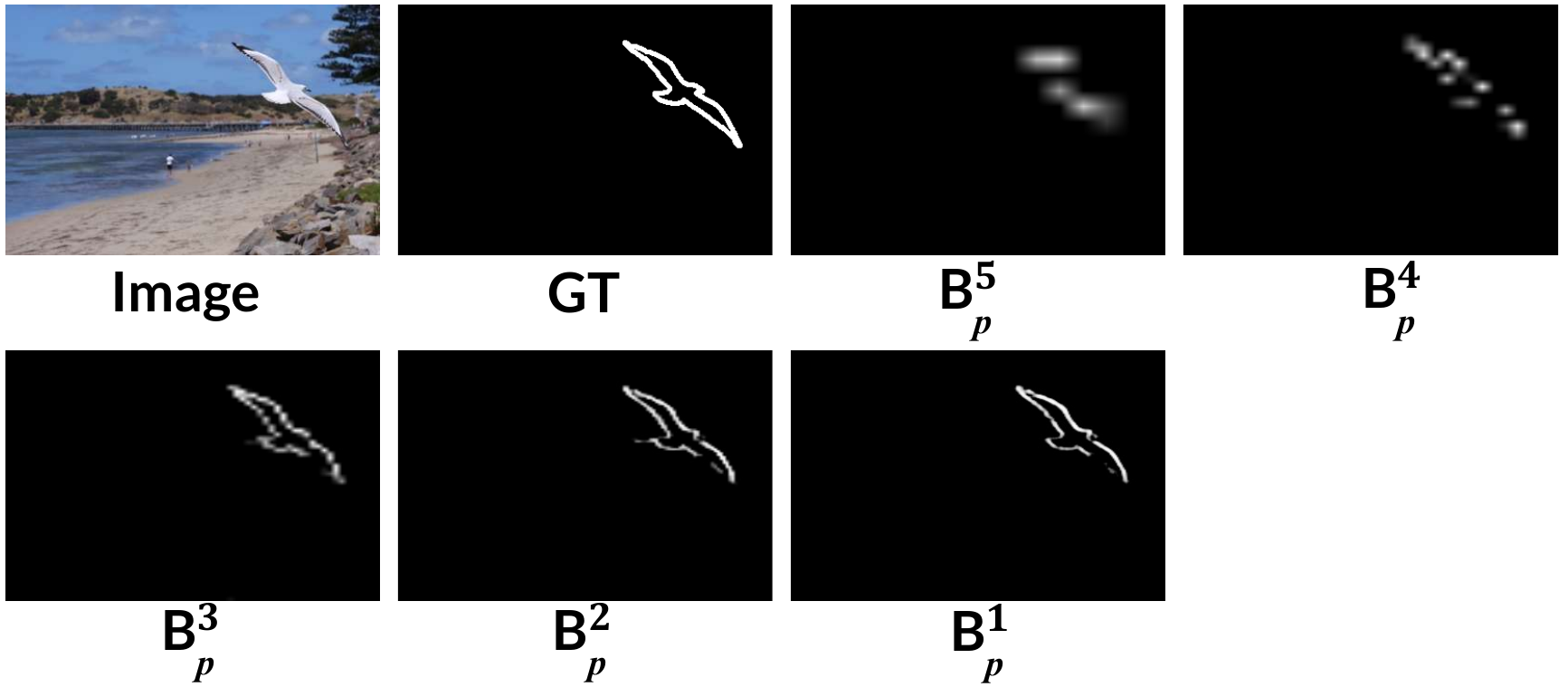} \ \\
\end{tabular}
\caption{Illustration of boundary ground truth and boundary prediction maps. $\bf{B}^{\tau}_{\it{p}} (\tau=1,2,...,5)$ represents the prediction results of level $\tau$. Note that we rescale the predictions to the same size for better visualization.}
\vspace{-6mm}
\label{boundary}
\end{figure}

To resolve the blurry boundary problem, we introduce the BPNet, which generates boundary predictions to guide saliency prediction.
To generate the boundary labels, we apply the open-source ``Canny” algorithm~\cite{Canny} to the binary saliency labels,  which usually provide in public saliency benchmarks. An example is shown in Fig.~\ref{boundary}.
The detailed structure of our BPNet is shown at the bottom of Fig.~\ref{diagram}.
We adopt five convolutional blocks of the VGG-16 model~\cite{simonyan2014very} to extract multi-scale boundary features.
Given an input image, our BPNet first extracts five scale feature maps $\bf{B}^{\tau}_{\it{f}}$.
To progressively merge multi-scale features and enlarge the boundary prediction map, we cascade several Residual Convolution Units (RCU)~\cite{Lin2017RefineNet} on the side-output feature maps.
%
In the scale $\tau$, BPNet generates boundary feature maps $\bf{B}^{\tau}$ by
\begin{equation}
\begin{aligned}
\bf{B}^{\tau}=
\begin{cases}
((\bf{W}^{\tau}{\star}_{s}\bf{B}^{{\tau}+1})\oplus \it{RCU}(\bf{B}_{\it{f}}^{\tau})),\rm{1}\le\it{\tau}<\rm{5}\\
\it{RCU}(\bf{B}_{\it{f}}^{\tau}),\it{\tau}=\rm{5}
\end{cases}
\label{boundary feature map}
\end{aligned}
\end{equation}
where $\bf{B}^{\it{\tau}}$ and $\bf{B}_{\it{f}}^{\it{\tau}}$ represent boundary feature maps and corresponding convolutional feature maps respectively.
${\star}_{s}$ denotes the deconvolution with a stride $s$ to ensure the same resolution.
$\oplus$ is the element-wise addition.
Then we apply a $1\times1$ convolution operation on the boundary feature maps of each scale to generate the boundary prediction map $\bf{B}^{\tau}_{\it{p}}$.
The comparison of boundary prediction maps is shown in Fig.~\ref{boundary}.
\vspace{-2mm}
\subsection{Attention-based Feature Fusion Module}
\begin{table*}[thp]
\begin{center}
\caption{Quantitative comparison on four large-scale datasets. The best two results are shown in \textcolor[rgb]{1,0,0}{red} and \textcolor[rgb]{0,0,1}{blue}, respectively.}
\begin{tabular}{|p{1.8cm}<{\centering}|p{1.3cm}<{\centering}|p{0.6cm}<{\centering} p{0.6cm}<{\centering} p{0.6cm}<{\centering} p{0.6cm}<{\centering}  p{0.6cm}<{\centering} p{0.6cm}<{\centering} p{0.6cm}<{\centering} p{0.6cm}<{\centering} p{0.6cm}<{\centering} p{0.6cm}<{\centering} p{0.6cm}<{\centering} p{0.6cm}<{\centering} p{0.6cm}<{\centering}|}
\hline
Dataset&Metric&Ours&Amulet&UCF&DHS&NLDF&RFCN&DS&DCL&ELD&LEGS&MDF&DRFI&BSCA\\
\hline
\multirow{3}{*}{ECSSD}&$\tt{F_{\beta}\uparrow}$&\textcolor[rgb]{1,0,0}{0.882}&0.867&0.839&0.872&\textcolor[rgb]{0,0,1}{0.878}&0.834&0.825&0.829&0.810&0.785&0.807&0.733&0.705\\
&$\tt{MAE}\downarrow$&\textcolor[rgb]{1,0,0}{0.051}&\textcolor[rgb]{0,0,1}{0.059}&0.078&0.059&0.063&0.107&0.122&0.088&0.079&0.118&0.105&0.164&0.182\\
\hline
\multirow{3}{*}{DUT-OMRON}&$\tt{F_{\beta}\uparrow}$&\textcolor[rgb]{1,0,0}{0.721}&0.669&0.613&/&\textcolor[rgb]{0,0,1}{0.683}&0.626&0.603&0.684&0.611&0.591&0.644&0.550&0.509\\
&$\tt{MAE}\downarrow$&\textcolor[rgb]{1,0,0}{0.060}&0.090&0.132&/&\textcolor[rgb]{0,0,1}{0.079}&0.111&0.120&0.097&0.092&0.133&0.092&0.139&0.190\\
\hline
\multirow{3}{*}{DUTS-TE}&$\tt{F_{\beta}\uparrow}$&\textcolor[rgb]{1,0,0}{0.763}&0.709&0.629&0.724&\textcolor[rgb]{0,0,1}{0.743}&0.712&0.632&0.714&0.628&0.585&0.673&0.541&0.499\\
&$\tt{MAE}\downarrow$&\textcolor[rgb]{1,0,0}{0.050}&0.080&0.117&0.067&\textcolor[rgb]{0,0,1}{0.065}&0.091&0.090&0.088&0.093&0.138&0.094&0.175&0.197\\
\hline
\multirow{3}{*}{HKU-IS}&$\tt{F_{\beta}\uparrow}$&\textcolor[rgb]{1,0,0}{0.887}&0.861&0.808&0.855&\textcolor[rgb]{0,0,1}{0.873}&0.835&0.785&0.853&0.769&0.723&0.801&0.722&0.654\\
&$\tt{MAE}\downarrow$&\textcolor[rgb]{1,0,0}{0.043}&0.053&0.074&0.053&\textcolor[rgb]{0,0,1}{0.048}&0.089&0.078&0.072&0.074&0.119&0.089&0.144&0.175\\
\hline
\end{tabular}
\label{table}
\vspace{-4mm}
\end{center}
\end{table*}
To efficiently utilize the boundary information and refine saliency maps, we introduce the attention-based feature fusion module (AFFM), which exploits multiple attention cues~\cite{Chen2017SCA}.
%
More specifically, we first reduce the channels of saliency features to the same size of boundary features by shrink dimension.
%
Then we perform a global average pooling on the saliency and boundary features to obtain a feature vector $\textbf{v}=[v_{1},v_{2},...,v_{n}]$ ($n$ is the channel dimension of layers).
A spatial softmax operator is applied on the feature vector to generate a normalized weight vector of the feature maps:
\begin{equation}
\begin{aligned}
{w_{i}}=\frac{e^{v_{i}}}{\sum_{i=1}^{n}e^{v_{i}}},
\label{boundary feature map}
\end{aligned}
\end{equation}
where $w_{i}$ is the weight of channel $i$ and $\sum_{i=1}^{n}w_{i}=1$. Subsequently, the fused feature maps $\bf{F}_{\text{fused}}^{\tau}$ is generated by
\begin{equation}
\begin{aligned}
\bf{F}_{\it{fused}}^{\tau}=(\bf{w}_{\it{F}}^{\tau}\otimes \bf{F}^{\tau}) \oplus (\bf{w}_{\it{B}}\otimes \bf{B}),
\label{equation8}
\end{aligned}
\end{equation}
where $\otimes$ is channel-wise product.
$\bf{B}$ denotes the boundary feature maps with the resolution of $256\times256$, which are in accordance with the aggregated saliency feature maps.
$\bf{w}_{\it{F}}^{\tau}$ and $\bf{w}_{\it{B}}$ represents the attention weights for saliency features and boundary features respectively.
With the fused feature, five paralleled fused prediction modules (FPM) (each of which is composed of a $3\times 3$ convolutional layer and an upsampling layer) are used to predict stage-wise prediction maps.
With the stage-wise prediction maps, we add another convolutional layer with a $1\times1$ kernel to predict the final prediction map.
\section{Experiments}
\label{sec:Experiments}
%
\subsection{Training and Testing Datasets}
\label{Dataset}
In the training process, the \textbf{DUTS-TR}~\cite{Lijun2017Learning} dataset is chosen as our training dataset, which includes 10, 553 images with accurate pixel-wise annotations.
We implement our proposed model based on the Caffe toolbox~\cite{jia2014caffe}.
We train and test our method with an NVIDIA 1080 GPU (with 8G memory).
%
%
The input image is uniformly resized into $256\times256\times3$ pixels and subtracted the ImageNet mean~\cite{deng2009imagenet}.
We find our model with this resolution achieves both effectiveness and efficiency.
We adopt the sigmoid cross entropy as the loss function for both saliency and boundary prediction.
Following previous works~\cite{Zhang2017Amulet,Zhang2017Learning,Zhang2018Salient}, we train the model until its training loss converges.
The weights of FCN backbones are initialized from the VGG-16~\cite{simonyan2014very} and ResNet-101~\cite{He2016Deep} models.
For other layers, we initialize the weights by the ``msra'' method.
We follow the parameters in~\cite{Zhang2017Amulet} and use the standard SGD method with a batch size 8, momentum 0.9 and weight decay 0.0005.
We set the learning rate to 1$\times$e$^{-8}$ and decrease it by 10\% after every 10 epoch.
%
%
Our model has a size of 410 MB and runs around 10 \emph{fps} for saliency inference, which is comparable even faster than most of methods.

We evaluate the performance of our method on four large-scale datasets described as follows.
\textbf{ECSSD}~\cite{Yang2013Saliency} is composed of 1000 images with random objects of different scales.
%
%
\textbf{HKU-IS}~\cite{Li2015Visual} includes 4447 images with fine pixel-wise annotations. Images of this dataset are
well chosen to include multiple disconnected salient objects or objects touching the image boundary.
\textbf{DUTS-TE}~\cite{Lijun2017Learning} has 5019 images with accurate pixel-wise annotations. All images are picked from the ImageNet DET test set and the SUN dataset~\cite{Xiao2010SUN}.
\textbf{DUT-OMRON}~\cite{Yang2013Saliency} has a total of 5168 high-quality images.
\begin{figure}[!t]
\centering
\begin{tabular}{@{}c@{}c@{}c@{}c}
\includegraphics[width=0.5\linewidth,height=2.6cm]{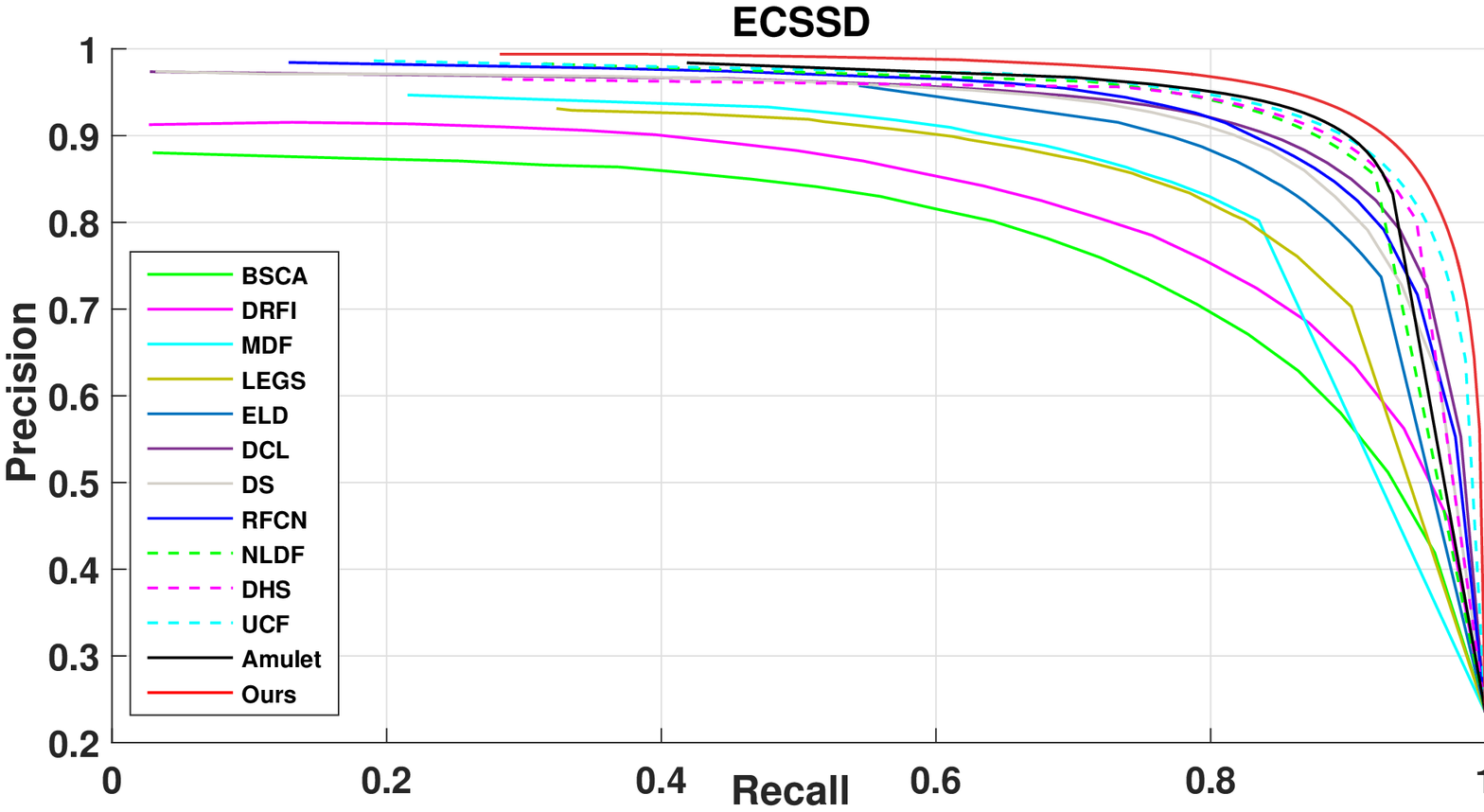}
\includegraphics[width=0.5\linewidth,height=2.6cm]{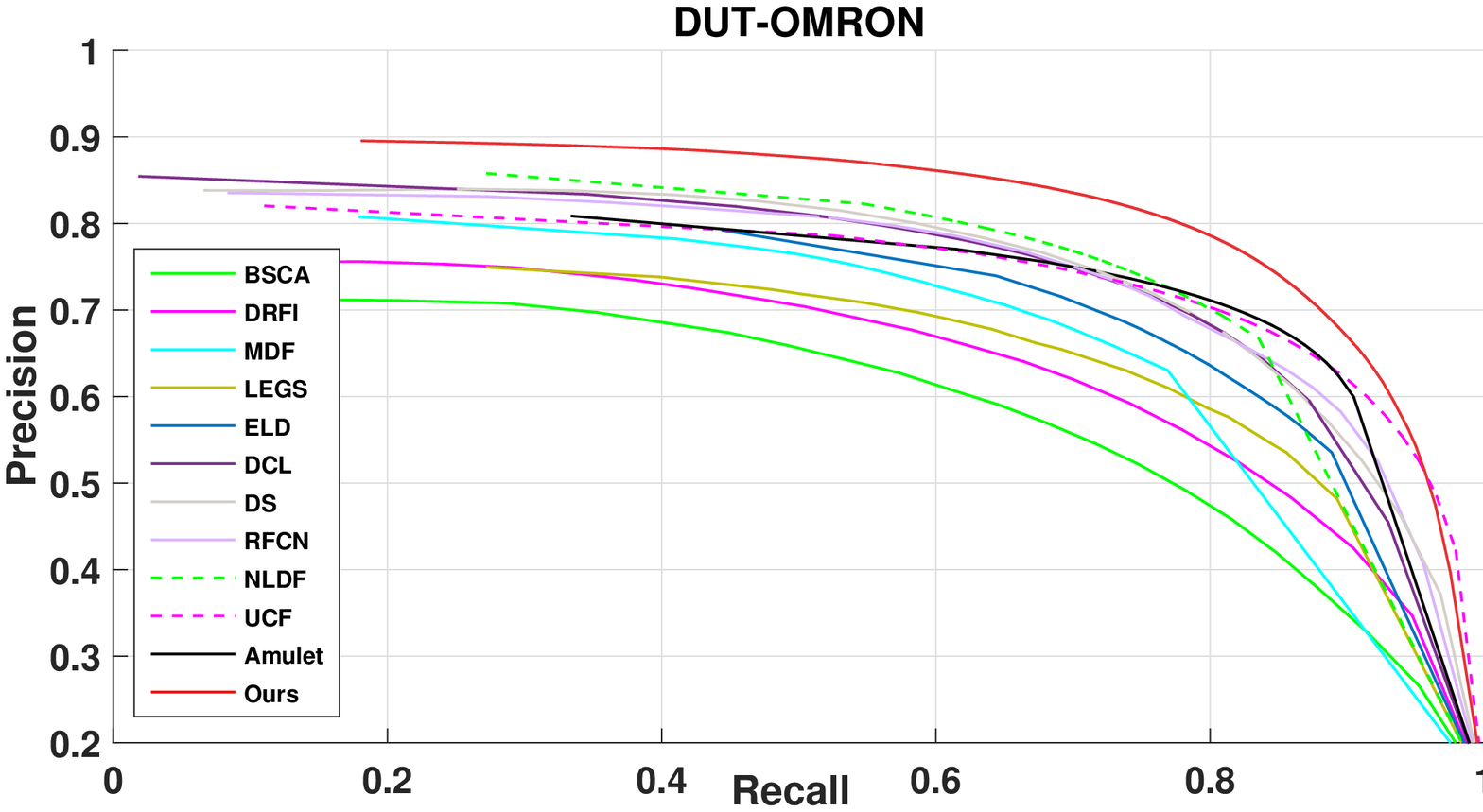}\\
\includegraphics[width=0.5\linewidth,height=2.6cm]{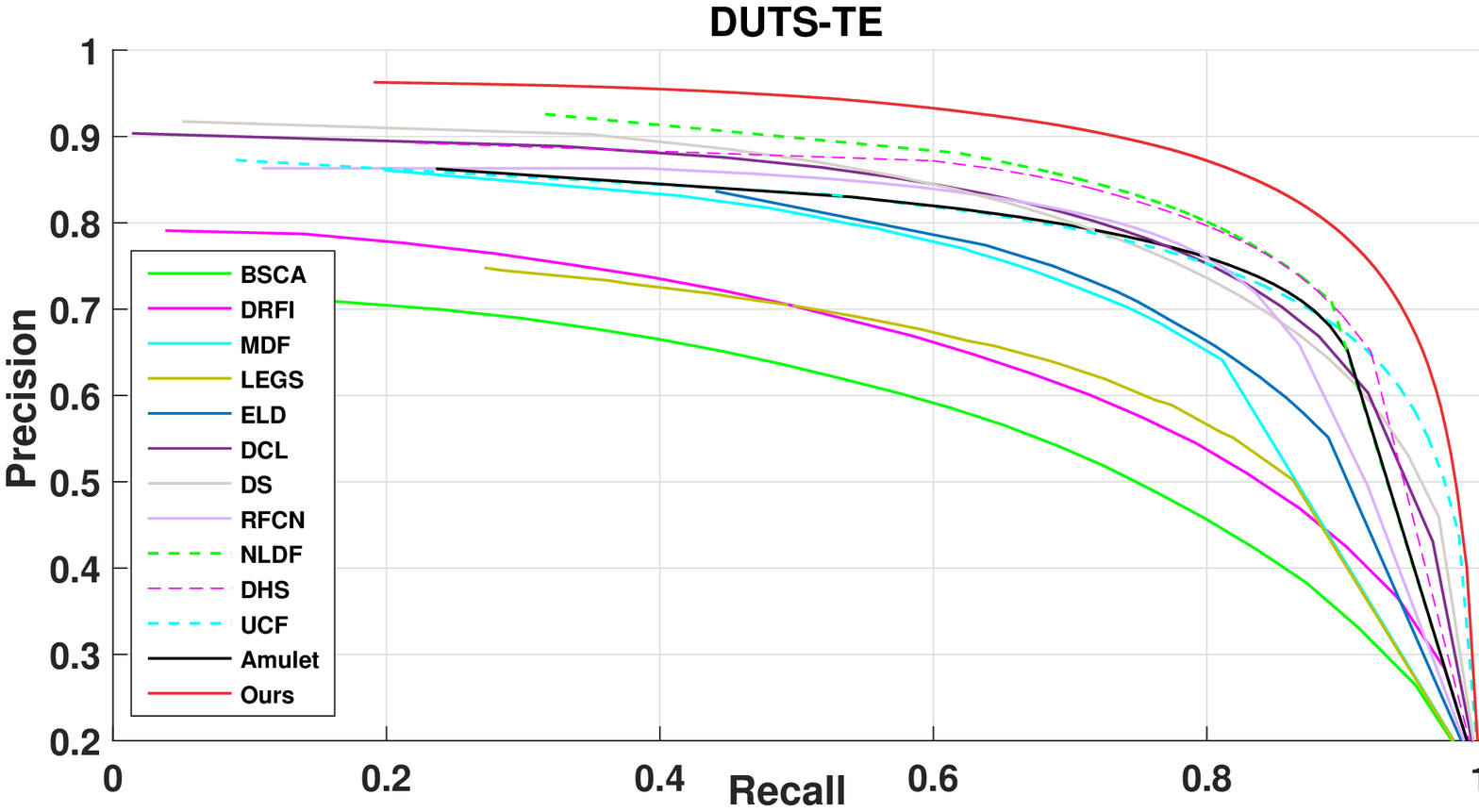}
\includegraphics[width=0.5\linewidth,height=2.6cm]{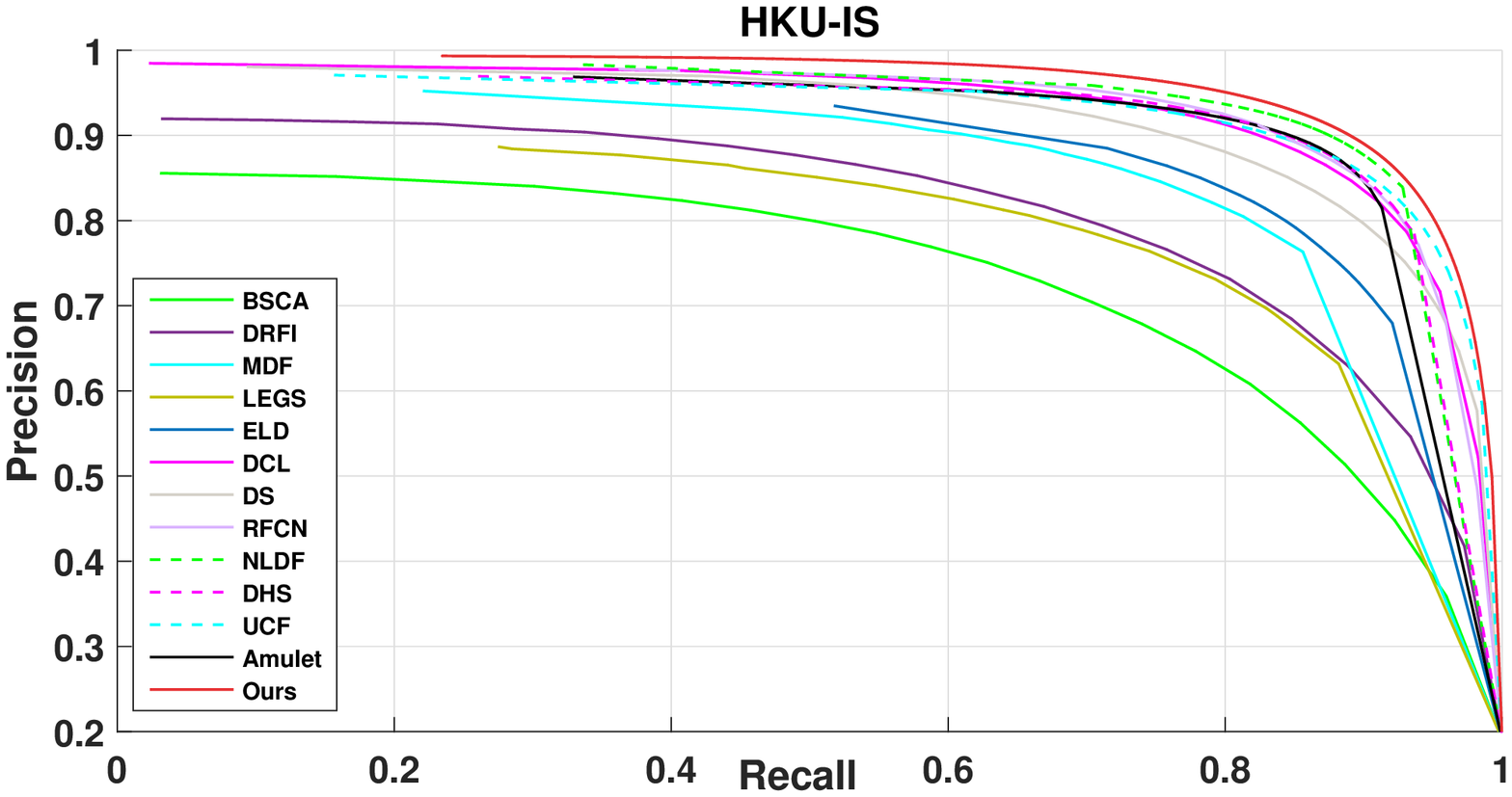}\\
\end{tabular}
 \caption{The PR curves of different state-of-the-art methods.
}
\vspace{-4mm}
\label{Fig pr}
\end{figure}
\vspace{-2mm}
\subsection{Evaluation Metrics.}
To evaluate the performance, we adopt three main metrics, \emph{i.e.}, the PR curves, mean F-measure score and mean absolute error (MAE)~\cite{Borji2015Salient}.
Precision-recall (PR) curves can be computed by binarizing the saliency map with a threshold in [0, 255] and then comparing the binary maps with the ground truth.
In many occasions, both precision and recall are important to measure methods.
Therefore, F-measure, which is averaged with precision and recall, is proposed to achieve the overall performance evaluation,
\begin{align}
F_{\beta} =\frac{(1+\beta^2)\times Precision\times Recall}{\beta^2\times Precision \times Recall}.
\end{align}
We set $\beta ^2$ to 0.3 to weigh precision more than recall as suggested in~\cite{Achanta2009Frequency,Borji2015What,Wang2015Deep}.
%

The above evaluations usually assign high saliency scores to salient pixels, which can be unfair especially for the methods which successfully detect non-salient regions, but miss the detection of salient regions.
Therefore, we also adopt the MAE metric~\cite{Borji2015Salient} to measure the average difference between the saliency prediction and the ground truth.
\begin{equation}
MAE =\frac{1}{W\times H}\sum_{x=1}^{W}\sum_{y=1}^{H}|S(x,y) - G(x,y)|,
\end{equation}
where $S$ is the predicted saliency map and $G$ is the binary ground truth mask.
It indicates how similar a saliency map is compared to the ground truth.
\vspace{-2mm}
\subsection{Comparison with Other Methods}
\begin{figure*}[!t]
\begin{center}
\begin{tabular}{@{}c}
\includegraphics[width=0.94\linewidth]{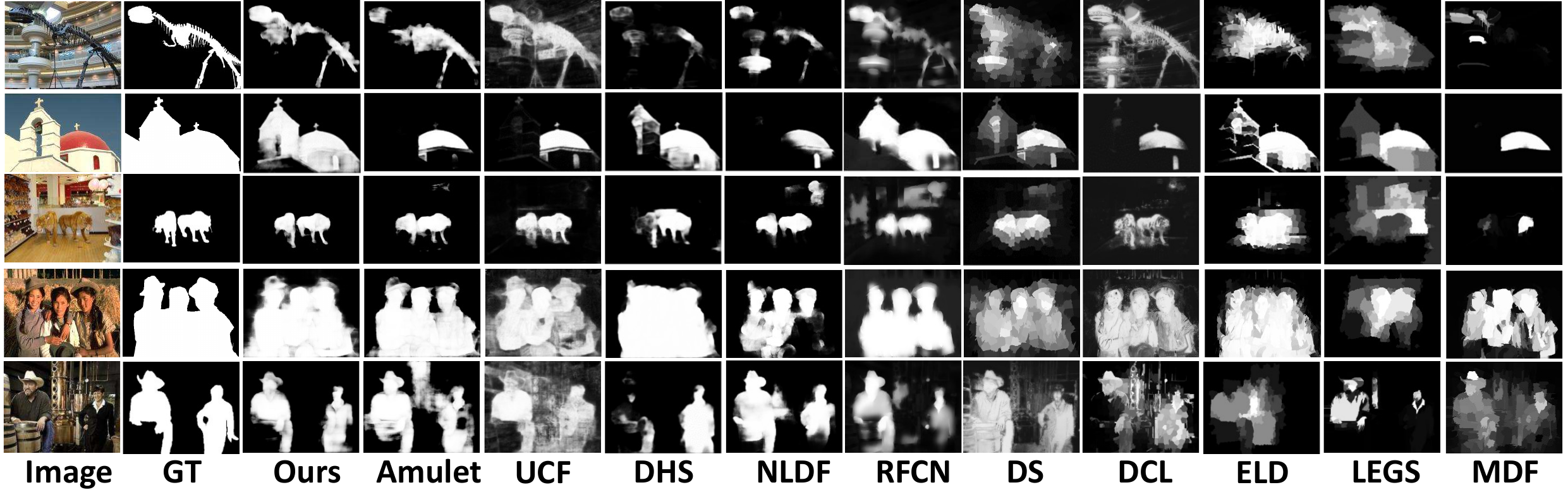}
\end{tabular}
{\caption{Comparisons of saliency maps with state-of-the-art methods. Due to the limitation of space, we do not show results of DRFI and BSCA methods.}
\label{Fig saliency}}
\vspace{-6mm}
\end{center}
\end{figure*}
We compared our method with other 12 algorithms, including 10 deep learning based algorithms
(Amulet\cite{Zhang2017Amulet}, UCF\cite{Zhang2017Learning}, NLDF\cite{Luo2017Non}, DHS\cite{Liu2016DHSNet}, DS\cite{Li2015DeepSaliency}, DCL\cite{Li2016Deep}, ELD\cite{Lee2016Deep}, RFCN\cite{Wang2016Saliency}, LEGS\cite{Wang2015Deep}, MDF\cite{Li2015Visual}) and 2 conventional  algorithms (DRFI\cite{Jiang2013Salient}, BSCA\cite{Qin2015Saliency}).
We compute saliency maps with the original implementations or use them provided by the authors.

\textbf{Quantitative Results}
As shown in Tab.~\ref{table} and Fig.~\ref{Fig pr}, our model consistently outperforms other methods across all the datasets in terms of all evaluation metrics, which convincingly demonstrates the effectiveness of the proposed method.
%

\textbf{Qualitative Results.}
Fig.~\ref{Fig saliency} shows visual comparison between our method and other algorithms.
As shown in the 1st row, the foreground is very complex, while our method successfully captures the main components against the competing algorithms.
The object in 2nd row is an unusual roof. Many algorithms fail to capture the semantic structure, while our method successfully highlights it. Our proposed method also perform better on images with similar color distribution between foreground and background (3rd row). For disconnected objects (last two rows), our method still performs well, while other algorithms misjudge the interferences.
\vspace{-2mm}
\subsection{Ablation Studies}
{\bf{Effects of the boundary branch.}} We verify the effectiveness of boundary branch in our framework.
The compared models include:
(1) With the same ResNet-101, we construct a baseline network, which is composed of encoder-decoder and feature combination structure~\cite{Zhang2017Amulet}.
The multi-level features are used to generate stage-wise prediction maps. The prediction map is generated by merging stage-wise prediction maps.
(2) We add boundary-stream to the baseline network and use concatenation operations to combine features of both branches for saliency detection.
The prediction scheme is similar with baseline network.
Different from (1), features of each level incorporate both saliency and boundary information. We name this setting as $Boundary^{+}$.
(3) To verify the effectiveness of cascaded RCUs, we implement the boundary-stream approach without RCUs, named $Boundary^{-}$
(4) Finally, AFFM is added to fuse saliency and boundary features to generate attentive features, resulting in our final model $AFFM^{+}$.

\begin{figure}[!t]
\begin{center}
\begin{tabular}{@{}c}
\includegraphics[width=0.9\linewidth,height=3cm]{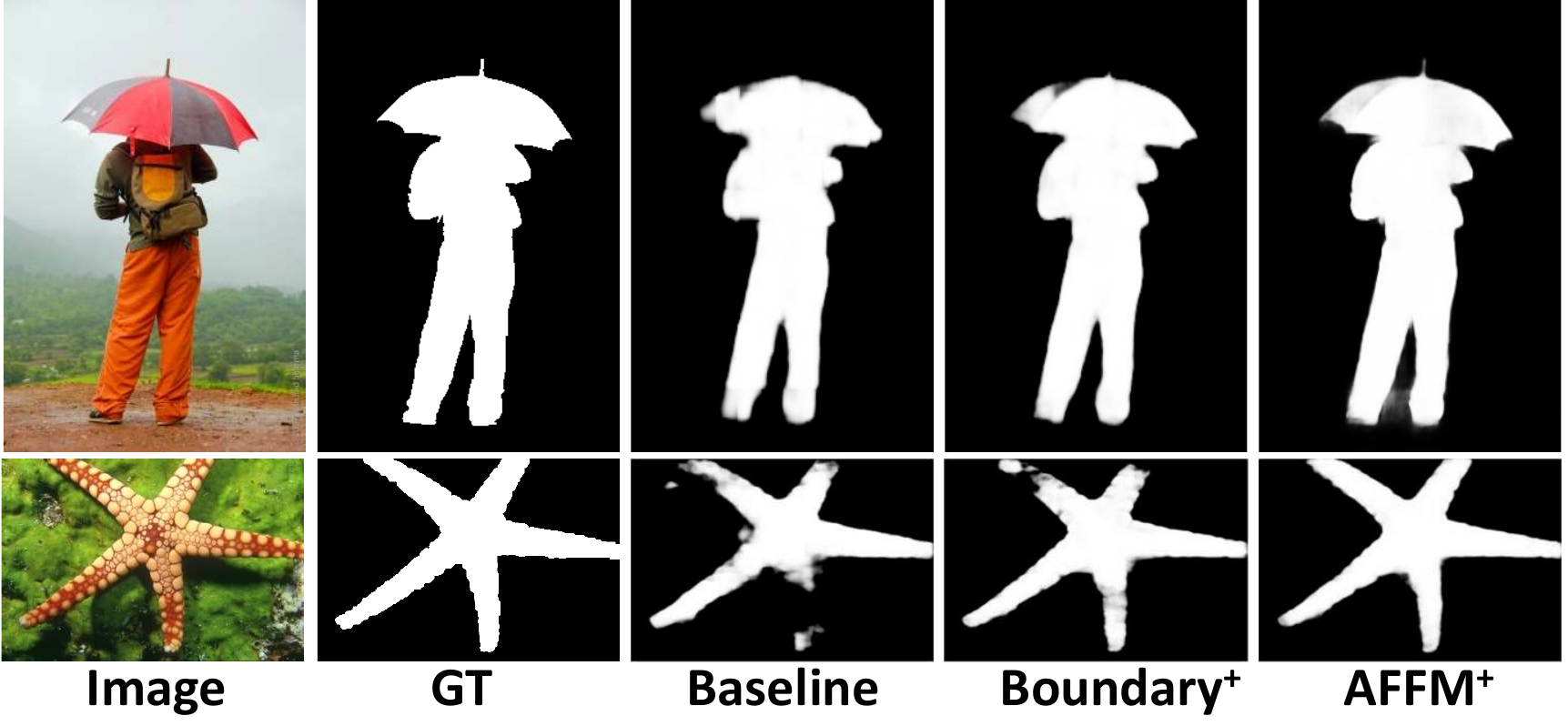}
\end{tabular}
{\caption{Qualitative comparisons with different model settings.}
\label{Fig_com}}
\vspace{-6mm}
\end{center}
\end{figure}

We perform detailed experiments on two datasets,~\emph{i.e.}, ECSSD and DUT-OMRON.
The results are shown in Tab.~\ref{table1}.
From quantitative evaluations and qualitative comparisons, one can observe that our proposed components effectively enhance saliency detection performance.
Especially, the boundary prediction module improves the MAE with 4\%.
Qualitative comparisons with different model settings are shown in Fig.~\ref{Fig_com}.
Compared with the baseline, our boundary-guided prediction is much better in predicting the boundary details.
The proposed method indeed produces saliency map with sharp boundaries. The low MAE metric also verifies this fact.
\begin{table}[thp]
\vspace{-2mm}
\begin{center}
\caption{Quantitative comparison of different settings. The best results are in bold. The VGG-16\cite{simonyan2014very} version is AFFM$^{+}\bullet$.}
\resizebox{0.45\textwidth}{!}
{
\begin{tabular}{|p{1.8cm}<{\centering}|p{1cm}<{\centering}|p{1cm}<{\centering}|p{1cm}<{\centering}|p{1cm}<{\centering}|}
\hline
\multirow{2}{*}{*}&\multicolumn{2}{c|}{ECSSD}&\multicolumn{2}{c|}{DUT-OMRON}\\
\cline{2-5}
&$\tt{F_{\beta}}\uparrow$&$\tt{MAE}\downarrow$&$\tt{F_{\beta}}\uparrow$&$\tt{MAE}\downarrow$\\
\hline
$Baseline$&0.8753&0.0568&0.7008&0.0761\\
$Boundary^{-}$&0.8712&0.0576&0.6944&0.0779\\
$Boundary^{+}$&$\normalsize{0.8797}$&0.0521&0.7083&0.0712\\
$AFFM^{+}$&$\normalsize{\bf{0.8821}}$&$\bf{0.0509}$&$\bf{0.7213}$&$\bf{0.0601}$\\
\hline
$AFFM^{+}\bullet$&0.8756&0.0558&0.7141&0.0669\\
\hline
\end{tabular}
}
\label{table1}
\vspace{-5mm}
\end{center}
\end{table}

\textbf{Effects of merging multi-scale predictions.} To verify the effects of the multi-scale FPM, we also perform a series experiments on ECSSD dataset.
Firstly, we evaluate the results of a single FPM.
%
Then we progressively add more FPMs to obtain the merged multi-scale results.
The quantitative results are listed in Tab.~\ref{table3}.
The ``$12345$” is our final model. From the results, we can observe that adding more FPMs can integrate more information, thus improving the performances.
\begin{table}[thp]
\begin{center}
\caption{Quantitative results of merging multi-scale predictions.}
\resizebox{0.5\textwidth}{!}
{
\begin{tabular}{|p{0.56cm}<{\centering}|p{0.56cm}<{\centering}|p{0.56cm}<{\centering}|p{0.56cm}<{\centering}|p{0.56cm}<{\centering}|p{0.56cm}<{\centering}| p{0.56cm}<{\centering}|p{0.56cm}<{\centering}|p{0.56cm}<{\centering}|p{0.56cm}<{\centering}|}
\hline
Metric&1&2&3&4&5&45&345&2345&12345\\
\hline
$\tt{F_{\beta}}\uparrow$&0.839&0.863&0.872&0.876&0.879&0.877&0.879&0.881&0.882\\
\hline
$\tt{MAE}\downarrow$&0.060&0.057&0.054&0.053&0.052&0.053&0.052&0.051&0.051\\
\hline
\end{tabular}
}
\label{table3}
\vspace{-6mm}
\end{center}
\end{table}
\section{Conclusion}
\label{sec:Conclusion}
In this paper, we propose a boundary-guided aggregating feature fusion network for salient object detection.
Different from the methods directly introduce high-level features into shallow layers, our method integrates feature maps into multiple resolutions.
The proposed attention-based feature fusion module can effectively refine the results with clear boundaries.


\ifCLASSOPTIONcaptionsoff
  \newpage
\fi



\bibliographystyle{IEEEtran}
\bibliography{IEEEabrv,refs}

%

\ignore{
\begin{IEEEbiography}{}

\end{IEEEbiography}

\begin{IEEEbiographynophoto}{John Doe}

\end{IEEEbiographynophoto}


\begin{IEEEbiographynophoto}{Jane Doe}
Biography text here.
\end{IEEEbiographynophoto}

}



\end{document}